%% file: main.tex
\documentclass{article}

    \PassOptionsToPackage{numbers, compress}{natbib}

    \usepackage[preprint]{neurips_2025}

\usepackage{microtype}
\usepackage{graphicx}
\usepackage[dvipsnames]{xcolor}
\usepackage{booktabs} 
\usepackage{bm}
\usepackage{multirow}
\usepackage{marvosym}
\usepackage{tcolorbox}
\usepackage{amsmath}
\usepackage{multicol}
\usepackage{colortbl}
\usepackage{color}
\usepackage{hyperref}
\usepackage[utf8]{inputenc} 
\usepackage[T1]{fontenc}    
\usepackage{hyperref}       
\usepackage{url}            
\usepackage{booktabs}       
\usepackage{amsfonts}       
\usepackage{nicefrac}       
\usepackage{microtype}      
\usepackage{xcolor}         

\usepackage{amssymb}
\usepackage{lipsum} 
\usepackage{fontawesome}
\usepackage{cleveref}
\usepackage{subcaption}
\usepackage{algorithm}
\usepackage{algpseudocode}
\usepackage{amsmath}
\usepackage{wrapfig}
\usepackage{multirow}

\title{Route to Reason: Adaptive Routing for LLM and Reasoning Strategy Selection}

\author{%
  Zhihong Pan, Kai Zhang, Yuze Zhao, Yupeng Han \\
  State Key Laboratory of Cognitive Intelligence \\
  University of Science and Technology of China\\
  \texttt{\{panzh,yuzezhao,yupenghan\}@mail.ustc.edu.cn} \\
  \texttt{kkzhang08@ustc.edu.cn} \\
}

\begin{document}

\maketitle

\input{sections/1_abstract}
\input{sections/2_introduction}
\input{sections/3_method}

\input{sections/4_experiment}

\input{sections/5_related_work}

\input{sections/6_discussion}

{\small
\bibliographystyle{IEEEtran}
\bibliography{main.bib}
}

\newpage
\input{sections/7_appendix}

\end{document}

%% file: sections/1_abstract.tex
\begin{abstract}
The inherent capabilities of a language model (LM) and the reasoning strategies it employs jointly determine its performance in reasoning tasks. While test-time scaling is regarded as an effective approach to tackling complex reasoning tasks, it incurs substantial computational costs and often leads to "overthinking", where models become trapped in "thought pitfalls". To address this challenge, we propose Route-To-Reason (RTR), a novel unified routing framework that dynamically allocates both LMs and reasoning strategies according to task difficulty under budget constraints. RTR learns compressed representations of both expert models and reasoning strategies, enabling their joint and adaptive selection at inference time. This method is low-cost, highly flexible, and can be seamlessly extended to arbitrary black-box or white-box models and strategies, achieving true plug-and-play functionality. Extensive experiments across seven open source models and four reasoning strategies demonstrate that RTR achieves an optimal trade-off between accuracy and computational efficiency among all baselines, achieving higher accuracy than the best single model while reducing token usage by over 60\%. The code is available at \href{https://github.com/goodmanpzh/Route-To-Reason}{https://github.com/goodmanpzh/Route-To-Reason}.

\end{abstract}

%% file: sections/2_introduction.tex
\section{Introduction}

With the continuous advancement of large language models (LLMs), their generality and autonomy have demonstrated human-like or even superhuman capabilities. In this context, reasoning ability has undoubtedly become the core driver of intelligent agent behavior~\citep{zhang2024codeagent}. Consequently, an increasing number of reasoning models~\citep{achiam2023gpt,yang2024qwen2,guo2025deepseek} and reasoning strategies~\citep{wang2022selfConsistency,besta2024graph,gao2023pal,xu2025chain,khot2022decomposed,chen2023program,cobbe2021gsm8k} have emerged. These expert models and reasoning strategies synergize and evolve, collectively pushing the boundaries of language models' reasoning capabilities.

This raises a critical question worthy of in-depth exploration: Given such a rich selection space of expert models and reasoning strategies, \textit{how can we efficiently identify the most suitable pairing within their combinatorial space?}

Intuitively, one might prefer combining powerful reasoning models (e.g., o3~\citep{openai2024o3o4}) with sophisticated reasoning strategies (e.g., Chain-of-Thought~\citep{wei2022chain}) to tackle complex problems. Particularly under the guidance of the test-time scaling paradigm, allocating a high budget to enhance performance appears to be a natural choice.

However, this intuition-driven, fixed pairing approach may face two key challenges in practice: \textbf{Performance bottlenecks}: Existing research~\citep{chen2024not,manvi2024adaptive,damani2025learning,liu2025efficient,li2025system,sui2025stop,yang2025towards,chen2025towards} suggests that ``overthinking'' can trap the reasoning process in protracted local reasoning patterns, limiting the model's ability to deviate from the current reasoning path, thereby degrading performance. \textbf{Budget inefficiency}: For low-difficulty tasks, employing high-performance models and complex strategies not only fails to yield significant gains but also leads to resource waste. We believe ``less is more'': lighter expert-strategy pairings often achieve better cost-performance trade-offs.

\begin{figure}[t]
    \centering
    \includegraphics[width=0.78\linewidth]{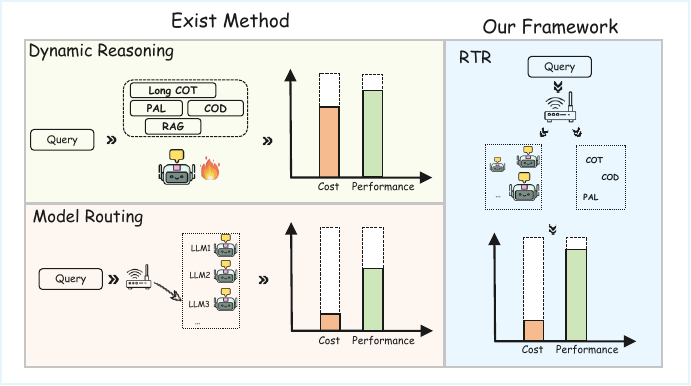}
    
    \caption{We propose Route-to-Reason (RTR), a low-cost and flexible expert selection framework capable of jointly optimizing model and strategy selection.}
    \label{model_difference}
    \vspace{-0.5cm}
\end{figure}

Some prior explorations have focused on model routing~\citep{zhuang2025embedllm,chen2024routerdc,feng2025graphrouter,mei2025eccos,wang2025mixllm,jitkrittum2025universal,hu2024routerbench,lu2024routing,ong2025routellm,chen2025harnessing}, enabling the system to select the most suitable model from a pool based on the input. However, most existing routing methods overlook the intricate relationship among expert model performance, reasoning strategies, and input complexity, often resulting in suboptimal decisions. While the works of ~\cite{damani2025learning,xu2025teaching,wang2024make,yue2024dots,parekh2024dynamic, aytes2025sketch} approach the problem from the perspective of dynamic reasoning strategy selection. These approaches, with the goal of tailoring the reasoning process to input characteristics, enhance performance and enable dynamic scaling at test time. Substantial performance variation across expert-strategy combinations and input difficulties remains underexplored. A principled approach to modeling these differences and selecting models accordingly could further improve performance and efficiency.

To address this, we propose a unified framework for joint model and strategy routing, enabling efficient and accurate test-time computation through dynamic selection. Specifically, we represent each expert and each reasoning strategy using learnable vector that capture their respective performance and computational cost characteristics. Given an input instance, we encode the query using a pretrained LM. Then, we design two modules to predict the expected performance and response tokens for all model-strategy combinations, thereby constructing a routing table. Based on this table, a routing policy selects the optimal model-strategy pair that maximizes efficiency while improving accuracy.

Compared to previous approaches, our framework dynamically adapts to questions of varying difficulty and intelligently selects the most appropriate model-strategy pair for each query. This leads to a more optimal trade-off between computational cost and reasoning performance. The key distinctions between our approach and existing methods are illustrated in Figure~\ref{model_difference}. Existing methods often fail to achieve an optimal balance between cost and performance. In contrast, by jointly selecting both the model and the reasoning strategy, our framework achieves superior performance at a reduced computational cost.

We conduct experiments on seven challenging reasoning tasks (language understanding, scientific reasoning and mathematical reasoning) to evaluate the proposed RTR in both in-distribution and out-of-distribution settings. Results show that our approach consistently improves reasoning accuracy while reducing the average number of generated tokens by over 60\% compared with single best model, validating its effectiveness.

%% file: sections/3_method.tex
\section{Methodology}

\subsection{Motivation}

\begin{figure}[t]
    \centering
    \vspace{-0.3cm}
    \begin{minipage}{0.48\linewidth}
        \includegraphics[width=\linewidth]{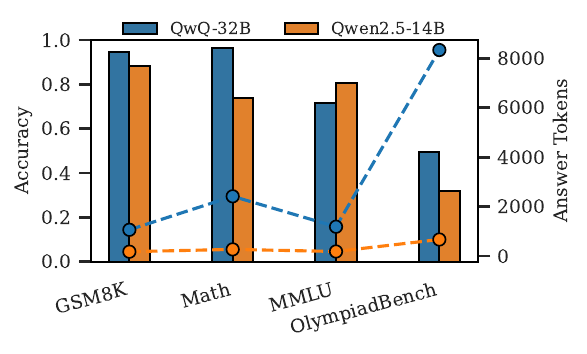}
        \caption{Performance and average answer tokens distribution of two different LLMs when responding to queries from subsets of four reasoning tasks.}
        \label{fig:motivation_1}
    \end{minipage}
     \hfill
    \begin{minipage}{0.48\linewidth}
        \includegraphics[width=\linewidth]{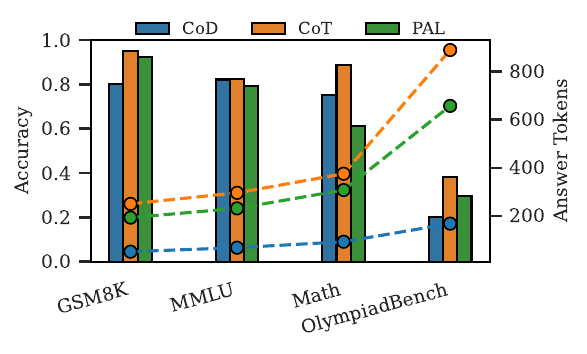}
        \caption{Performance and average answer tokens distribution of Qwen2.5-14B-Instruct under different reasoning strategies when responding to queries from subsets of four reasoning tasks.}
        \label{fig:motivation_2}
    \end{minipage}
    \vspace{-0.2cm}
\end{figure}

Recently, reasoning-enhanced language models have demonstrated strong performance on complex tasks by leveraging extended reasoning steps and structured thinking~\citep{yang2024qwen2,guo2025deepseek}. However, their advantages tend to diminish on simpler problems. As illustrated in Figure~\ref{fig:motivation_1}, while the thinking model (QwQ) shows substantial improvements over the non-thinking model (Qwen2.5-14B-Instruct) on complex tasks such as Math~\citep{math} and OlympiadBench~\citep{he2024olympiadbench}, it offers only marginal improvements or even slight performance drops on relatively straightforward and commonsense tasks like GSM8K~\citep{cobbe2021gsm8k} and MMLU~\citep{mmlu}, despite incurring significantly higher inference costs, often generating up to 10 times more tokens. This highlights a critical challenge: rather than universally deploying heavy reasoning models, it becomes essential to determine when such complex reasoning is truly necessary.

Moreover, we observe that reasoning strategies play as crucial a role as model selection. As shown in Figure~\ref{fig:motivation_2}, strategies such as CoD~\citep{xu2025chain} prompting can significantly reduce answer length while achieving performance comparable to CoT~\citep{wei2022chain} prompting on easier tasks. Additionally, PAL~\citep{gao2023pal} performs well on arithmetic-intensive tasks and generates more consistent outputs. These observations motivate our approach: jointly selecting both the model and the reasoning strategy for each input query enables adaptive, cost-effective inference while maintaining strong overall performance.

\subsection{Problem Formulation and Preliminaries}

We consider a collection of language models $\mathcal{M} = \{m_j : j = 1, \dots, M\}$, each differing in size or capability, and a set of reasoning strategies $\mathcal{S} = \{s_k : k = 1, \dots, K\}$. Given a set of input queries $\mathcal{D} = \{x_i : i = 1, \dots, N\}$, applying model $m_j$ with strategy $s_k$ to query $x_i$ yields a response characterized by two quantities: the performance score $a_{i,j,k}$ (e.g., accuracy, utility, or another task-specific metric), and the number of generated tokens $l_{i,j,k}$ serving as a proxy for inference cost.
Our objective is to predict these two quantities and select an appropriate model-strategy pair $(m_j, s_k)$ based on these quantities for each query $x_i$.
Formally, we seek to learn a routing function:
\[
\pi: \mathcal{X} \rightarrow \mathcal{M} \times \mathcal{S},
\]
where $\pi(x_i) = (j^*, k^*)$ denotes the selected model and strategy for $x_i$. The objective is to optimize the trade-off between total performance and total generation cost:
\[
\max_{\pi} \sum_{i=1}^n a_{i, j^*, k^*} - \lambda \sum_{i=1}^n l_{i, j^*, k^*},
\]
where $\lambda > 0$ is a hyperparameter that balances performance and efficiency, $a_{i, j^*, k^*}$ and $l_{i, j^*, k^*}$ denote the performance and generated length, respectively.

\subsection{RTR Framework}

\begin{figure*}[t]
    \centering
    \vspace{-0.5cm}
    \includegraphics[width=\linewidth]{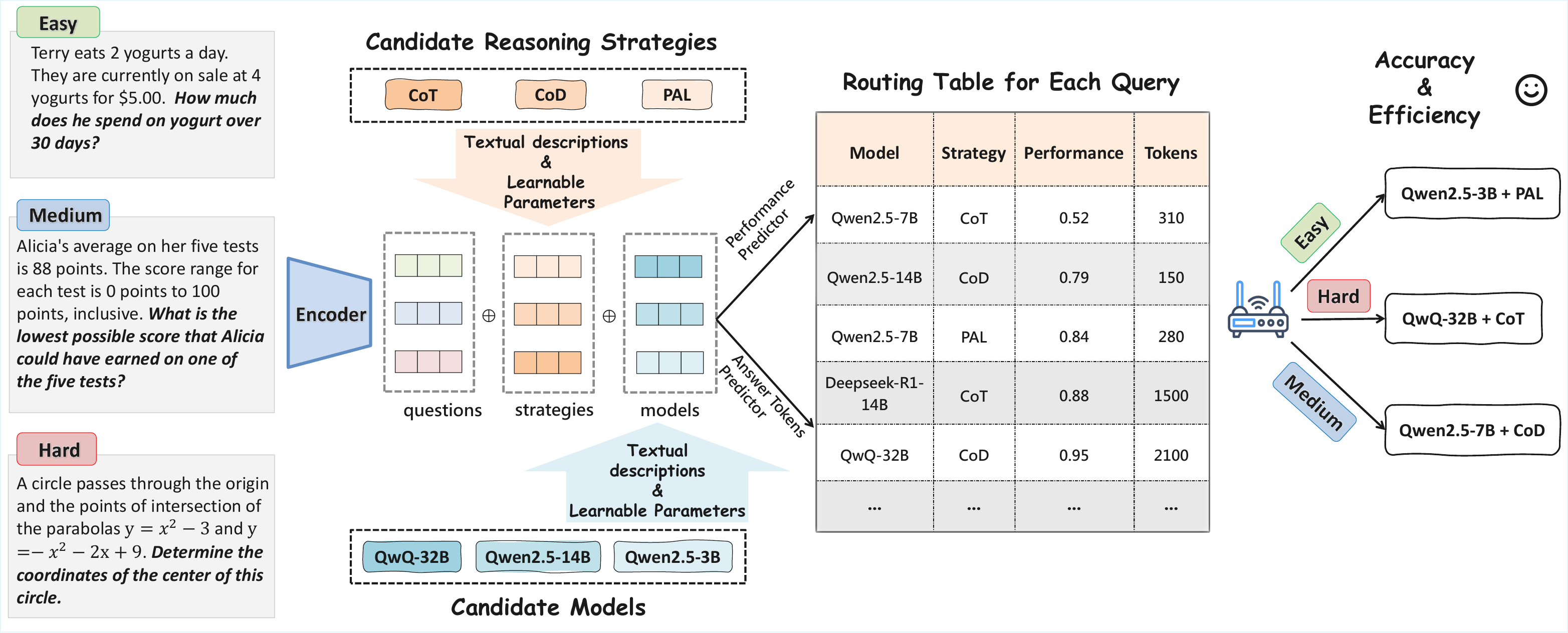}
    \vspace{-0.1cm}
    \caption{
        The RTR first encodes the input question, available models, and reasoning strategies. Two predictor modules then estimate the expected performance and token usage for each model-strategy combination, generating a routing table. Finally, the router selects the most suitable model-strategy pair that balances accuracy and efficiency for each question.
    }
    \label{fig:rtr_pipeline}
    \vspace{-0.2cm}
\end{figure*}

\paragraph{Method Overview.}
As shown in Figure~\ref{fig:rtr_pipeline}, for each input question, we first encode it into a dense vector representation using a pretrained encoder. Additionally, each candidate model and reasoning strategy is represented by an embedding that captures its characteristics, such as performance capability and computational cost. These embeddings, together with the question representation, are concatenated and passed into two predictors: one estimates the expected performance, and the other predicts the expected answer length. This yields a routing table for each question containing the predicted performance and token usage for every model-strategy pair. Finally, a routing policy, controlled by a trade-off coefficient between performance and cost, selects the optimal model and strategy combination for each query. The training and inference procedures are provided in Algorithm~\ref{alg:rtr}.

\paragraph{Model and Strategy Representation.}
The prediction performance for each model-strategy pair critically depends on how well their characteristics are represented. To this end, we design a dual-component representation for each model and strategy. The first component is obtained by encoding a textual description of the model or strategy using a pretrained encoder, similar to how we encode input questions. These descriptions are generated using an auxiliary LLM aiming to capture high-level semantic attributes such as capability and usage scenarios. The detailed textual descriptions used for models and strategies are provided in Appendix~\ref{sec:model_strategy_description}. This component is fixed during training. The second component is a randomly initialized embedding that is learned and updated during training. This learnable component enables the model to capture fine-grained behavioral nuances and task-specific adaptation signals that may not be easily conveyed through textual descriptions. These two components combine external knowledge and internal task-specific patterns to form a comprehensive representation.

\paragraph{Dual Prediction Module.}
After obtaining the representations of the query, model, and strategy, we construct two predictor modules to estimate the response quality and output token usage of each model-strategy combination for a given query. Specifically, we use multi-layer perceptron (MLP) for each prediction task. These predictors are trained independently using their respective loss functions.

The performance prediction module estimates the likelihood that a given model-strategy pair will correctly answer the input query. Since the correctness labels in the training data are binary, we use a binary cross-entropy loss:
\begin{equation}
    \hat{a}_{i,j,k} = \mathrm{MLP}_{\mathrm{perf}}(z_{i,j,k}), \quad \mathcal{L}_{\mathrm{perf}} = - y \log(\hat{a}_{i,j,k}) - (1 - y)\log(1 - \hat{a}_{i,j,k}),
\end{equation}
where $z_{i,j,k}$ denotes the input to the performance predictor for query $i$ and model-strategy pair $(j, k)$, and $y$ is the ground-truth correctness label.

The token usage prediction module estimates the expected number of output tokens required to generate a response. As this is a regression task, we adopt mean squared error (MSE) loss:
\begin{equation}
    \hat{l}_{i,j,k} = \mathrm{MLP}_{\mathrm{len}}(z_{i,j,k}), \quad \mathcal{L}_{\mathrm{len}} = \left( \hat{l}_{i,j,k} - l_{i,j,k} \right)^2,
\end{equation}
where $z_{i,j,k}$ denotes the input to the token usage predictor, and $l_{i,j,k}$ is the ground-truth response length.

\paragraph{Routing Table and Routing Policy.}
Inspired by routing concepts in computer networks, we design a routing mechanism based on a predicted performance-cost table. For each query, we generate a routing table that organizes the predicted performance and output tokens for all combinations of candidate models and strategies as shown in Figure~\ref{fig:rtr_pipeline}. This table encapsulates the necessary decision-making information, akin to a routing table in network systems, and serves as the foundation for downstream routing decisions.

To determine the best routing destination for a given query, we propose a simple yet effective scoring function. The final routing score $s_{n,l,s}$ for query $q_n$ routed to model $l$ with strategy $s$ is computed as a weighted combination of predicted performance and cost:

\begin{equation}
    \text{score}_{i,j,k} = \lambda \cdot \hat{a}_{i,j,k} - (1 - \lambda) \cdot \hat{l}_{i,j,k}.
\end{equation}

Here, $\hat{a}_{i,j,k}$ and $\hat{l}_{i,j,k}$ denote the predicted performance score and predicted cost (e.g., based on output length) respectively. The parameter $\lambda \in [0,1]$ controls the trade-off between quality and cost: a higher $\lambda$ prioritizes performance (quality-first), while a lower $\lambda$ encourages cost-efficiency. The optimal routing decision is then made by selecting the candidate with the highest score:

\begin{equation}
    (j^*, k^*) = \arg\max_{j, k} \; \text{score}_{i,j,k}.
\end{equation}

This approach offers flexibility in adapting to different application needs. For instance, by tuning $\lambda$, users can customize the system to prefer higher-quality responses, lower costs, or a balanced trade-off between the two. We do experiments to show its effectiveness.

\begin{wrapfigure}{r}{0.5\textwidth}
\vspace{-1.8cm}
\begin{minipage}{0.48\textwidth}
\begin{algorithm}[H]

\caption{Training and Inference of RTR.}
\label{alg:rtr}
\begin{algorithmic}[1]
\Require Training set $\mathcal{D}_\text{train}$, encoder $E$, descriptions $\{d_j, d_k\}$, trade-off $\lambda$
\State Initialize learnable embeddings $\{e_j\}, \{e_k\}$ and MLPs: $\text{MLP}_\text{perf}$, $\text{MLP}_\text{len}$
\State Encode descriptions: $z_j = [E(d_j); e_j]$, $z_k = [E(d_k); e_k]$
\For{each $(x_i, \{a_{ijk}, l_{ijk}\}_{j,k})$ in $\mathcal{D}_\text{train}$}
    \State $q_i = E(x_i)$
    \For{each $(j, k)$}
        \State $z_{i,j,k} = [q_i; z_j; z_k]$
        \State Update $\text{MLP}_\text{perf}$, $\text{MLP}_\text{len}$ using BCE and MSE losses
    \EndFor
\EndFor
\State \textbf{Inference:}
\For{each test $x_n$}
    \State $q_n = E(x_n)$
    \For{each $(j, k)$}
        \State $z_{n,j,k} = [q_n; z_j; z_k]$
        \State Predict $\hat{a}_{n,j,k}$, $\hat{l}_{n,j,k}$ via MLPs
        \State $\text{score}_{n,j,k} = \lambda \hat{a}_{n,j,k} - (1-\lambda)\hat{l}_{n,j,k}$
    \EndFor
    \State Select $(j^*, k^*) = \arg\max_{j,k} \text{score}_{n,j,k}$
\EndFor
\end{algorithmic}

\end{algorithm}
\end{minipage}
\vspace{-1cm}
\end{wrapfigure}

%% file: sections/4_experiment.tex
\section{Experiment}
\subsection{Experimental Setup} \label{experimental_setup}

\paragraph{Candidate LLMs and Strategies.}\label{sec:llm_strategies}
We choose six open-source LLMs from HuggingFace\footnote{https://huggingface.co/}:
Non-thinking models\citep{yang2024qwen2} include Qwen2.5-3B, Qwen2.5-7B, and Qwen2.5-14B, which are general-purpose models designed for a wide range of tasks, with performance generally improving with model size. Thinking models \citep{yang2024qwen2,guo2025deepseek} include DeepSeek-R1-7B, DeepSeek-R1-14B, and QwQ-32B, which are optimized for reasoning tasks. These models are trained to produce longer reasoning steps for complex problems.

We evaluate four reasoning strategies: \textbf{Vanilla}, which uses the original question without any additional prompting and serves as a baseline; \textbf{CoT}~\citep{wei2022chain} prompts the model to generate intermediate reasoning steps before answering; \textbf{PAL}~\citep{gao2023pal} prompts the model to solve questions by generating executable code; and \textbf{CoD}~\citep{xu2025chain} prompts the model to generate only intermediate drafts with explicit constraints on output length, encouraging concise reasoning. Detailed prompts for each strategy are provided in Appendix \ref{sec:prompts}.

\paragraph{Datasets.}

The statistics of the datasets used are summarized in Table~\ref{tab:datasets}. We select subsets from four reasoning tasks of varying difficulty: (i) GSM8K~\citep{cobbe2021gsm8k}, a mathematical reasoning dataset with diverse grade school word problems; (ii) MMLU~\citep{mmlu}, a general-purpose benchmark, where we evaluate on a selected subset of STEM subjects; (iii) Math~\citep{math}, a large-scale dataset focusing on diverse math problems requiring numerical reasoning and problem solving; and (iv) OlympiadBench~\citep{he2024olympiadbench}, a challenging benchmark derived from olympiad-level scientific problems.  
We construct our dataset by collecting responses generated by candidate LLMs using different reasoning strategies, and split it into 70\% for training and 30\% for testing.

To assess generalization, we further evaluate the trained router on three out-of-distribution (OOD) benchmarks: (i) SciQ~\citep{SciQ}, a multiple-choice science question dataset for middle school level; (ii) PIQA~\citep{bisk2020piqa}, which tests physical commonsense reasoning through everyday scenarios; and (iii) ARC-C~\citep{allenai:arc}, a challenging subset of the ARC benchmark focusing on questions requiring multi-step reasoning and commonsense inference.  
These datasets are not included in the training phase and are used solely to evaluate the router’s out-of-distribution performance.

\begin{wraptable}[12]{r}{0.53\textwidth}
\vspace{-0.6cm}
\centering
\caption{Overview of Datasets.}
\label{tab:datasets}
\setlength{\tabcolsep}{4.5pt}
\renewcommand{\arraystretch}{1.1}
\begin{tabular}{l l c c}
\toprule
 & Dataset & Cases & Difficulty \\
\midrule
\multirow{4}{*}{ID}
    & GSM8K~\citep{cobbe2021gsm8k}          & 600 & Easy \\
    & Math~\citep{math}            & 600 & Medium \\
    & MMLU~\citep{mmlu}            & 600 & Medium \\
    & OlympiadBench~\citep{he2024olympiadbench}   & 300 & Hard \\
\midrule
\multirow{3}{*}{OOD}
    & SciQ~\citep{SciQ}         & 300 & Easy \\
    &  PIQA~\citep{bisk2020piqa}        & 300 & Medium \\
    & ARC-C~\citep{allenai:arc}         & 300 & Hard \\
\bottomrule
\end{tabular}
\end{wraptable}

\paragraph{Baselines.}
We compare our proposed model against the following baselines. First, we consider two single-model baselines: \textbf{Qwen2.5-3B}: The smallest single LLM in our pool of candidate LLMs; \textbf{QwQ-32B}: The best single LLM in our pool of candidate LLMs.

Next, we compare our method with several routing-based baselines. The detailed settings of these baselines are provided in Appendix~\ref{sec:baseline}.
\textbf{Random}: Randomly selects a model-strategy pair from the pool of candidates; \textbf{KNN-Router}~\citep{hu2024routerbench}: For a given test query, the KNN-router identifies the $k$ nearest neighbors from the training set in the embedding space. For each candidate model-strategy pair, it estimates the expected performance by averaging the observed rewards over these neighbors, and selects the pair with the highest estimated score; \textbf{RouteLLM}~\citep{ong2025routellm}: A binary router that assigns queries to either a strong or weak model based on learned routing policies over query features; \textbf{EmbedLLM}~\citep{zhuang2025embedllm}: Constructs an embedding for each model using matrix factorization to capture its performance profile across the dataset, and uses these embeddings to route queries to the most suitable LLM.

\paragraph{Evaluation metrics.}
Routing methods are evaluated using two primary metrics: \textit{accuracy}, which measures the correctness of the selected model's response on each task, and the \textit{average number of tokens}, reflecting the efficiency in controlling computational cost.

\paragraph{Implementation Details.} For non-reasoning models, outputs are generated via greedy decoding (temperature = 0). For reasoning-based models, we follow official recommendations, using temperature = 0.6 and top-p = 0.95 for sampling. For the router, textual descriptions of each model, reasoning strategy, and input query are encoded with the \texttt{all-mpnet-base-v2}\footnote{\url{https://huggingface.co/sentence-transformers/all-mpnet-base-v2}} model from the sentence-transformers library, yielding 768-dimensional embeddings. Both predictors are implemented as MLPs with a hidden size of 768. Training is performed for up to 100 epochs using the Adam optimizer (batch size 32, initial learning rate $1 \times 10^{-3}$), with early stopping based on test loss.

\subsection{Main Results}

\paragraph{In-Distribution Results.}
Table~\ref{tab:routing-results} presents the performance comparison of various routing methods across four reasoning tasks. First, we observe that all routing models outperform random routing, demonstrating the effectiveness of employing routing strategies. Our proposed method, \textbf{RTR}, achieves the best overall accuracy (82.5\%) while significantly reducing the average output length (1091.3 tokens). It outperforms all baselines in terms of the trade-off between performance and cost. Specifically, RTR achieves the highest accuracy on MMLU and OlympiadBench, and the second-best results on GSM8K and Math. Compared to the best-performing baseline, EmbedLLM, RTR matches or surpasses its accuracy while reducing the average token usage by over 39.6\%. Notably, compared to the single largest model, QwQ-32B, which achieves strong performance at a very high cost, RTR improves average accuracy by 2.5 percentage points while reducing output token length by more than 60\%. These results demonstrate that RTR can effectively select the most appropriate LLM and strategy to achieve both high accuracy and efficiency.

\paragraph{Out-Of-Distribution Results.}
As shown in Table~\ref{tab:ood_results}, the proposed RTR achieves the highest average accuracy across all out-of-distribution datasets, surpassing the best-performing individual LLM (QwQ) by a substantial margin of 0.5\%. Notably, RTR achieves this strong performance while significantly reducing the average number of response tokens (393.9 vs. 1387.3), indicating superior efficiency. Among all routing-based baselines, only RTR is able to outperform QwQ in terms of average accuracy, highlighting its better generalization capability.

\begin{table*}
\Large
\vspace{-0.5cm}
\centering
\caption{Results on in-distribution datasets. The best router-based result is in \textbf{bold}, and the second-best is \underline{underlined}.}
\resizebox{1.0\linewidth}{!}{ 
\begin{tabular}{lcccccccc|cc}
\toprule
\multirow{2}{*}{Model} & \multicolumn{2}{c}{GSM8K} & \multicolumn{2}{c}{Math} & \multicolumn{2}{c}{MMLU} & \multicolumn{2}{c}{OlympiadBench} & \multicolumn{2}{|c}{AVG} \\
\cmidrule(lr){2-3} \cmidrule(lr){4-5} \cmidrule(lr){6-7} \cmidrule(lr){8-9} \cmidrule(lr){10-11}
& Accuracy & Tokens & Accuracy & Tokens & Accuracy & Tokens & Accuracy & Tokens & Accuracy & Tokens \\
\midrule
Qwen2.5-3B & 71.5 & 205.4 & 54.3 & 295.5 & 61.7 & 253.1 & 21.0 & 1007.2 & 56.0 & 371.7 \\
QwQ & 94.4 & 1148.5 & 95.6 & 2583.0 & 71.8 & 1219.3 & 48.5 & 8762.3 & 80.0 & 2745.2 \\
\midrule
random & 84.4 & 382.4 & 77.7 & \underline{1043.7} & 69.4 & 723.1 & 32.6 & 4219 .1& 69.5 & 1271.6 \\
KNN-Router & 89.2 & \textbf{272.9} & 88.3 & 1122.3 & 78.1 & \textbf{347.5} & 36.6 & \underline{4197.7} & 76.9 & \underline{1101.3} \\
RouteLLM & 91.0 & 372.0 & 89.3 & 1161.8 & 72.8 & 597.3 & \textbf{45.5} & 7696.8 & 77.3 & 1814.3 \\
EmbedLLM & \textbf{95.8} & 927.1 & \textbf{94.8} & 1898.2 & \underline{80.5} & 508.8 & 41.5 & 5786.4 & \underline{81.9} & 1808.3 \\
\midrule
\rowcolor{gray!15}
\textbf{RTR (Ours)} & \underline{95.2} & \underline{297.8} & \underline{92.9} & \textbf{982.9} & \textbf{82.5} & \underline{432.5} & \textbf{45.5} & \textbf{3399.7} & \textbf{82.5} &  \textbf{1091.3} \\
\bottomrule
\end{tabular}
}
\label{tab:routing-results}
\end{table*}

\begin{table*}
\Large
\centering
\caption{Results on out-of-distribution datasets. The best is in \textbf{bold} and the second-best is \underline{underlined}.}
\resizebox{0.8\linewidth}{!}{ 
\begin{tabular}{lcccccc|cc}
\toprule
\multirow{2}{*}{Model} & \multicolumn{2}{c}{PIQA} & \multicolumn{2}{c}{SciQ} & \multicolumn{2}{c}{ARC-C} & \multicolumn{2}{|c}{AVG} \\
\cmidrule(lr){2-3} \cmidrule(lr){4-5} \cmidrule(lr){6-7} \cmidrule(lr){8-9}
& Accuracy & Tokens & Accuracy & Tokens & Accuracy & Tokens & Accuracy & Tokens \\
\midrule
Qwen2.5-3B & 73.3 & 150.8 & 68.2 & 187.6 & 65.3 & 248.9 & 68.9 & 195.8 \\
QwQ & 95.5 & 1126.4 & 93.8 & 1203.6 & 91.7 & 1831.9 & 93.7 & 1387.3 \\
\midrule
random & 84.6 & 350.0 & 78.3 & 430.8 & 74.1 & 713.9 & 79.0 & 498.2 \\
KNN-Router & 90.3 & \underline{277.4} & 89.2 & \textbf{376.2} & 86.3 & \textbf{550.6} & 88.6 & \underline{419.4} \\
RouteLLM & 93.2 & 363.1 & \underline{92.7} & 894.3 & 91.2 & 1103.2 & 92.3 & 786.9 \\
EmbedLLM & \underline{95.1} & 832.8 & 92.2 & 1002.3 & \underline{92.4} & 1631.1 & \underline{93.2} & 1155.4 \\
\midrule
\rowcolor{gray!15}
\textbf{RTR (Ours)} & \textbf{95.3} & \textbf{222.3} & \textbf{94.2} & \underline{405.7} & \textbf{93.1} & \underline{553.7} & \textbf{94.2} & \textbf{393.9} \\
\bottomrule
\end{tabular}
}
\label{tab:ood_results}
\end{table*}

\subsection{Further Analysis}

\begin{figure}[b]
    \centering
    \vspace{-0.5cm}
    \begin{minipage}{0.45\linewidth}
        \includegraphics[width=\linewidth]{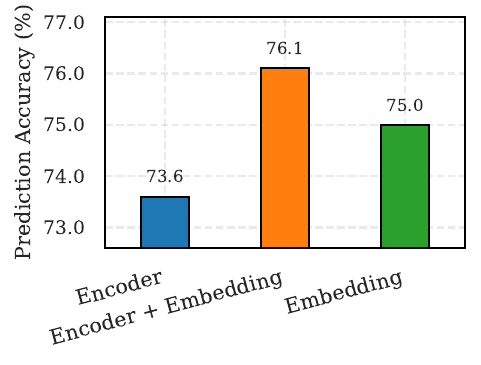}
        \caption{Distribution of the performance and average answer token of different LLMs in response to queries on the 4 reasoning tasks.}
        \label{fig:prediction_acc}
    \end{minipage}
     \hfill
    \begin{minipage}{0.45\linewidth}
        \includegraphics[width=\linewidth]{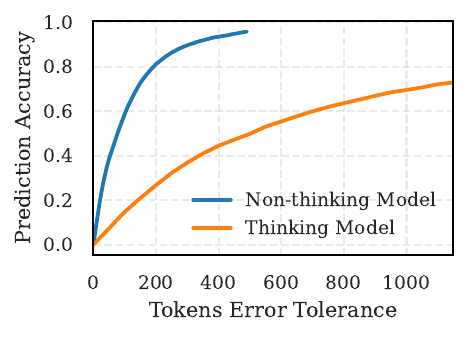}
        \caption{Distribution of the performance and average answer token of different reasoning strategies responding to queries on 4 reasoning tasks.}
        \label{fig:prediction_length}
    \end{minipage}
\end{figure}

\paragraph{Effectiveness of Performance Prediction.}
We conduct ablation studies to evaluate the effectiveness of our dual-component representation for model-strategy pairs. We compare three configurations: (1) textual descriptions encoded using a sentence encoder, (2) randomly initialized learnable embeddings, and (3) a combination of both. As shown in Figure~\ref{fig:prediction_acc}, the combined representation yields the highest accuracy in predicting model-strategy correctness, demonstrating the complementary strengths of prior knowledge from textual descriptions and task-adaptive learned embeddings.

\paragraph{Effectiveness of Token Usage Prediction.}
We evaluate the performance of our token length prediction. As shown in Figure~\ref{fig:prediction_length}, although reasoning and non-reasoning models differ substantially in average output length, the predictor achieves approximately 80\% accuracy for non-reasoning models within a 200-token error margin, and about 60\% accuracy for reasoning models with a 600-token margin. Given that the output length of reasoning models can exceed 10,000 tokens, this level of coarse-grained prediction accuracy is sufficient for effective routing.

\begin{figure}[htbp]
  \centering
  \begin{minipage}[t]{0.48\textwidth}
    \centering
    \includegraphics[width=\linewidth]{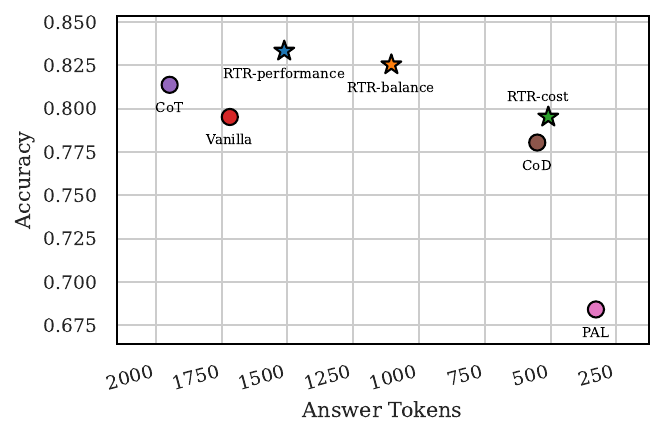}
    \caption{
      Distribution of different routing strategies. Routing with any of the four single strategies often results in suboptimal performance. All three routing configurations outperform the Vanilla baseline.
    }
    \label{fig:methods_tradeoff}
  \end{minipage}
  \hfill
  \begin{minipage}[t]{0.48\textwidth}
    \centering
    \includegraphics[width=\linewidth]{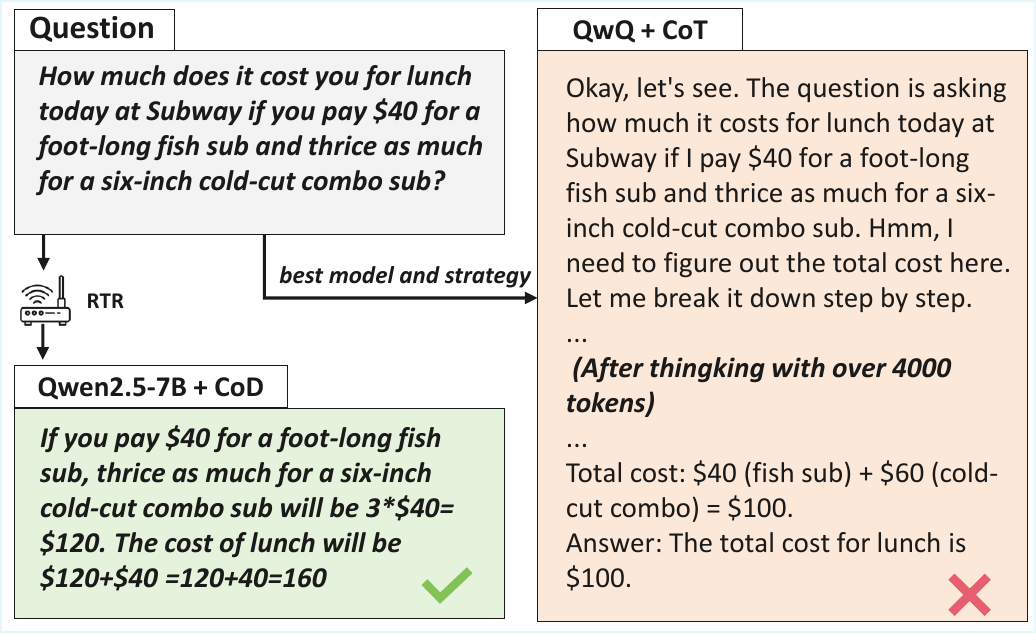}
    \caption{
    In this case, without using RTR, selecting the best model (QwQ) and the CoT strategy leads to redundant reasoning and ultimately an incorrect answer. In contrast, our RTR routes the query to Qwen2.5-7B with the CoD strategy and obtains the correct answer using only 32 tokens.
    }
    \label{fig:case}
  \end{minipage}
  \vspace{-0.5cm}
\end{figure}

\paragraph{Different Routing Strategy.}
We further evaluate our model under various routing scenarios. We first consider routing based on a single fixed strategy. As shown in Figure~\ref{fig:methods_tradeoff}, relying on a single strategy often leads to higher costs or suboptimal performance, as it cannot leverage the complementary strengths of different strategies. For example, strategies with greater reasoning depth may achieve higher accuracy but incur significantly higher costs. This underscores the advantage of dynamically selecting reasoning strategies.

We then examine three routing configurations by adjusting the controller parameters to reflect different trade-off preferences: \textit{performance-first}, \textit{cost-first}, and \textit{balanced}, corresponding to RTR-performance, RTR-cost, and RTR-balanced in Figure~\ref{fig:methods_tradeoff}, respectively. Experimental results show that our framework is flexible under different budget constraints and achieves competitive performance while adapting to varying priorities in cost-effectiveness and accuracy. Moreover, all three configurations outperform using only the vanilla strategy, highlighting the benefits of leveraging multiple reasoning strategies.

\subsection{Case Study}
Figure~\ref{fig:case} illustrates a case showing the effectiveness of RTR. When RTR is disabled, the best-performing model (QwQ) along with the reasoning strategy of Chain-of-Thought results in redundant and unnecessarily verbose reasoning, ultimately leading to an incorrect answer. 
In contrast, when RTR is applied, it routes the query to Qwen2.5-7B and CoD. This configuration yields the correct answer while using only 32 tokens, demonstrating both improved accuracy and significantly reduced computation.

\subsection{When To Think Routing}

\begin{wraptable}{r}{0.47\textwidth}
\vspace{-1.4cm}
\caption{
Performance of dynamic routing between reasoning and non-reasoning modes in Qwen3-4B.
}
\label{tab:qwen3}
\centering
\setlength{\tabcolsep}{5pt}
\renewcommand\arraystretch{1.2}
\resizebox{0.4\textwidth}{!}{
\begin{tabular}{@{}l c r@{}}
\toprule
Method & Acc (\%) & Tokens \\
\midrule
Qwen3-4B (non-thinking)        & 73.4  & 592.1 \\
Qwen3-4B (thinking)       & 82.4  & 3112.8 \\
\midrule
Random &    76.6  & 1834.2 \\
RouteLLM     & 82.8  & 2247.1 \\
KNN-Router     & 80.6  & 1418.7 \\
EmbedLLM    & 83.2  & 2623.5 \\
\midrule
\rowcolor{gray!15}
\textbf{RTR (ours)} & \textbf{83.8}  & \textbf{1321.1} \\
\bottomrule
\end{tabular}%
}
\vspace{-0.5cm}
\end{wraptable}

With the emergence of recent LLMs that support manual toggling between \textit{thinking} and \textit{non-thinking} modes, our framework is naturally compatible with such dual-mode models via binary routing. Specifically, we represent each mode as a distinct candidate in our model pool, this setup enables our router to automatically determine when to invoke the thinking mode, allowing for dynamic selection based on task requirements. To validate this capability, we conduct experiments with Qwen3-4B, which supports both reasoning and non-reasoning modes. As shown in Table~\ref{tab:qwen3}, our framework effectively learns to trigger the reasoning mode only when beneficial, thereby significantly reducing average token usage while maintaining high prediction accuracy. Moreover, dynamic strategy selection improves performance beyond that of any single-model baseline.

%% file: sections/5_related_work.tex
\section{Related Work}
\paragraph{Efficient LLM Reasoning.}
Recently, LLMs~\citep{yang2024qwen2,guo2025deepseek,hurst2024gpt} have demonstrated increasingly powerful reasoning capabilities, but this progress has come at the expense of significantly higher inference-time computation. Recent studies have shown that such improvements often result in the problem of overthinking~\citep{chen2024not,manvi2024adaptive,damani2025learning,liu2025efficient,li2025system,sui2025stop,yang2025towards,chen2025towards}, where models tend to generate unnecessarily long and redundant reasoning steps, especially for simple queries. This has brought growing attention to the challenge of dynamically adapting inference based on problem difficulty. Existing approaches typically address this by fine-tuning~\citep{pan2024dynathink,kang2025c3ot,manvi2024adaptive} the model or applying reinforcement learning~\citep{chen2024not,sui2025stop,yang2025towards} to shorten reasoning trajectories and reduce unnecessary token generation, thereby enabling adaptive computation during inference.

\paragraph{Dynamic Reasoning  Strategies in LLM.}
Chain-of-Thought (CoT) prompting~\citep{wei2022chain} and its variants, such as decomposition-based methods~\citep{zhao2025unveiling,khot2022decomposed} and Tree-of-Thought (ToT)~\citep{yao2024tree}, have significantly improved performance across a wide range of reasoning tasks. Recent studies~\citep{yue2024dots,sprague2025to,parekh2024dynamic} have shown that these strategies exhibit complementary strengths depending on the characteristics of the task. For instance, while CoT performs well on mathematical and logical reasoning, it may not be universally effective for all question types. The Program-aided Language (PAL) approach~\citep{gao2023pal} has shown superior performance on arithmetic-intensive tasks by incorporating code execution into the reasoning process. Recent work~\citep{yue2024dots,parekh2024dynamic,aytes2025sketch} has studied how to select the most suitable reasoning strategy for different types of questions.

\paragraph{LLM Routing.}
Model routing has emerged as one of the most cost-effective approaches in LLM ensembles~\citep{feng2025graphrouter,chen2025harnessing,ding2024hybrid,hu2024routerbench}, as it selects a single model to generate responses for individual queries. RouteLLM~\citep{ong2025routellm} proposes four distinct strategies for routing between small and large models, effectively reducing cost. EmbedLLM~\citep{zhuang2025embedllm} proposes learning universal model embeddings to facilitate routing decisions across diverse models. RouterBench~\citep{hu2024routerbench} introduces a benchmark dataset for routing tasks and implements a range of routing baselines to balance response quality and computational cost. GraphRouter~\citep{feng2025graphrouter} adopts a graph-based approach by representing queries, tasks, and models as nodes in a heterogeneous graph to jointly predict model performance and cost. Recent works~\citep{mei2025eccos,wang2025mixllm,jitkrittum2025universal} have further explored collaborative multi-model deployments and the development of general-purpose routing systems. However, most existing approaches fail to consider the interplay between reasoning strategies and model performance. Our work addresses this gap by enabling more efficient inference and improved performance through joint reasoning strategy and model selection.

%% file: sections/6_discussion.tex
\section{Discussion}
\label{sec-conclusion}

\paragraph{Conclusion.}

In this paper, we present \textbf{RTR}, a unified framework for jointly selecting the optimal model and reasoning strategy for each query. To the best of our knowledge, RTR is the first approach to simultaneously address both model and strategy selection in a single routing framework. 
By representing all candidate models and strategies as learnable embeddings, our framework predicts two key metrics for each query: the expected performance score and the answer token usage across all available model-strategy pairs. These predictions are used to construct a routing table, enabling RTR to dynamically determine the best combination of model and reasoning strategy for each query.
Extensive experiments on four in-distribution and three out-of distribution datasets demonstrate that RTR consistently makes effective routing decisions, achieving a favorable trade-off between overall performance and computational cost across all of the baselines. Furthermore, RTR is capable of dynamically deciding when to invoke the reasoning mode in models with binary thinking capabilities.

\paragraph{Limitations and Future Directions.}
Our current evaluation focuses on single-model selection, although incorporating collaborative decision-making across multiple models represents a promising avenue for future work. Additionally, while our experiments are centered on reasoning tasks, extending evaluation to a broader range of problem domains would further validate the framework’s generality. We believe that efficient and intelligent routing systems are crucial for the future of LLMs, particularly in multi-model and resource-constrained scenarios.

%% file: sections/7_appendix.tex
\appendix

\newpage
\section{Extended Experiment Results}

\subsection{Effectiveness of Multi-Strategy and Multi-Model Selection}

\begin{figure*}[htbp]
    \centering
    \includegraphics[width=0.85\linewidth]{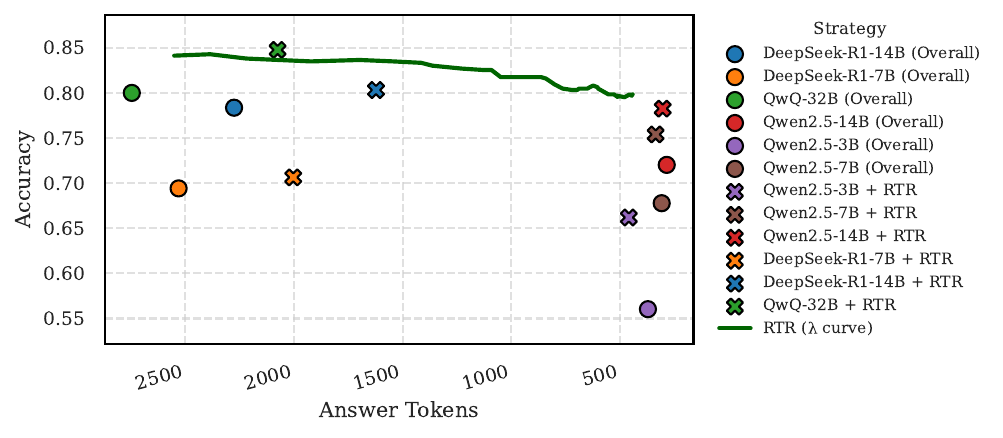}
    \caption{
    Effectiveness of Multi-Model and Multi-Strategy.
    }
    \label{fig:model_rtr_curve}
\end{figure*}

\paragraph{Effectiveness of Multi-Strategy.} We evaluate the benefits of multi-strategy selection by comparing the original accuracy and the routing accuracy of our proposed RTR to those of individual models. As shown in Figure~\ref{fig:model_rtr_curve}, we compare using RTR within a single model (labeled as ‘Model + RTR’, marked by $\times$) to the model’s original overall accuracy (labeled as ‘Model Overall’, marked by $\circ$). Routing across multiple strategies consistently improves performance across all models, highlighting the value of strategy selection tailored to instance characteristics. 

\paragraph{Effectiveness of Multi-Model.} As shown in Figure~\ref{fig:model_rtr_curve}, we plot the $\lambda$-curve ($\lambda$ controls the trade-off between token usage and performance) to evaluate the effectiveness of multi-model selection. While RTR’s $\lambda$-curve shows only a slight performance drop compared to routing on QwQ alone, it significantly outperforms most individual models. This demonstrates that leveraging multiple models with strategy-aware routing effectively balances accuracy and efficiency.

\subsection{Effectiveness of Training Samples}

\begin{figure*}[b]
    \centering
    \includegraphics[width=0.6\linewidth]{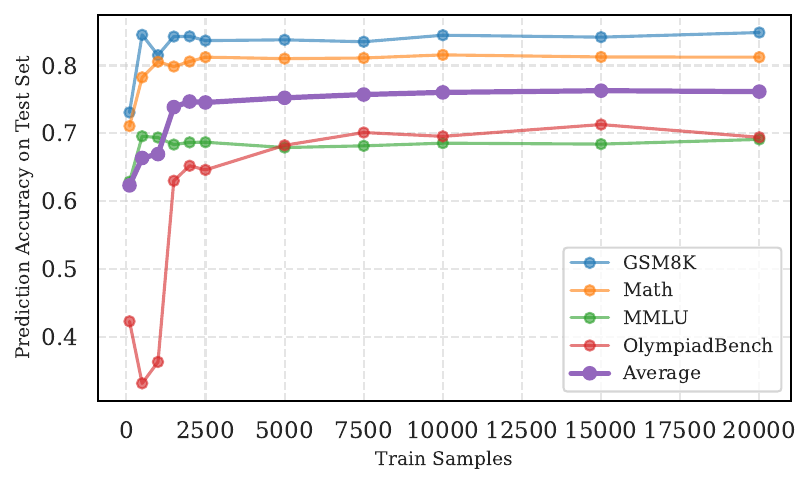}
    \caption{
    Prediction accuracy on the test set under different training sample sizes.
    }
    \label{fig:samples_num}
\end{figure*}

We evaluate the prediction performance of the performance predictor under varying training sample sizes. As shown in Figure~\ref{fig:samples_num}, an increase in the number of training samples leads to a general improvement in prediction accuracy across all datasets. This enhancement is particularly pronounced when the training sample size is relatively small. As the training size approaches around 5000 samples, the performance begins to converge. This suggests that our prediction framework is capable of effectively modeling the relationship between model-strategy pairs and problems using a relatively small amount of data, thereby enabling accurate performance prediction.

\subsection{More Case Studies}
We show several cases of RTR in Figures~\ref{fig:case1},~\ref{fig:case2}, and~\ref{fig:case3} to illustrate its ability to adaptively select model-strategy pairs. As demonstrated in Figure~\ref{fig:case1}, when faced with relatively simple problems, RTR selects a non-reasoning model and a lightweight strategy, which also leads to a correct answer with a lower token cost. In Figures~\ref{fig:case2} and~\ref{fig:case3}, when handling harder questions, RTR benefits from accurate performance and token cost prediction, ensuring both correctness and lower token usage. This demonstrates the effectiveness of RTR in model-strategy pair routing.
\input{cases/case1}
\input{cases/case2}
\input{cases/case3}

\section{Additional Details on Methodology and Experiments}
\subsection{Baseline Implementation Details}\label{sec:baseline}

We describe the implementation details of the baseline methods as follows.
For the \textbf{Random} router, a model-strategy pair is randomly selected for each input. We report the average performance over 50 runs.
For the \textbf{KNN-Router}, we set the parameter $k$ to 10. For each model-strategy pair, we compute its performance over all neighboring queries, where the score is defined as the average accuracy minus the average output length multiplied by a parameter $\lambda$. We tune $\lambda$ to balance cost and performance, and report the best result.
For \textbf{RouteLLM}, since this is a binary router, we label each query according to its average accuracy on the candidate models, and train a binary classifier to decide which model to route each query to. We tune the balance parameter and report the best result. The strong and weak models used are QwQ and Qwen2.5-7B, respectively.
For \textbf{EmbedLLM}, we assign a unique ID to each model-strategy pair and learn a 768-dimensional embedding for each pair. Following the original method, we frame the learning process as a reconstruction task: given a matrix of model correctness across prompts, a reconstruction network is trained to recover this matrix, thereby enforcing the embeddings to capture the behavioral characteristics of each model-strategy pair.

\subsection{Prompts Used in Experiments}\label{sec:prompts}

\begin{tcolorbox}
\textbf{CoT}\\
Please reason step by step before providing the final answer, and put your final answer within \verb|\\boxed{{}}|.\\

\textbf{CoD}\\
Think step by step, but only keep minimum draft for each thinking step, with 5 words at most, and put your final answer within \verb|\\boxed{{}}|.\\

\textbf{PAL}\\
Write a Python code snippet to solve the following problem. Do not use any plotting libraries or the input() function.
\end{tcolorbox}

\subsection{Profiles of Strategies and Models} \label{sec:model_strategy_description}
\begin{tcolorbox}
\textbf{Vanilla:} Vanilla prompting retains the original question content without adding any additional prompt information.\\

\textbf{Chain-of-Thought (CoT):} Chain-of-Thought (CoT) prompting guides the model to articulate a step-by-step reasoning process before providing the final answer. This results in longer responses and slower reasoning speed, typically generating the longest answers, but it performs best on complex problems such as mathematical reasoning.\\

\textbf{Chain-of-Draft (CoD):} Chain-of-Draft (CoD) prompts the model to generate only intermediate drafts with explicit constraints on output length, encouraging concise reasoning. These drafts represent the model's thinking process, often containing important calculation steps and key reasoning information. It simplifies the intermediate steps of the reasoning chain while retaining good performance, resulting in shorter answers.\\

\textbf{Program-Aided Language (PAL):} Program-Aided Language (PAL) transforms the reasoning process into executable code. This approach leverages the determinism of programming languages to ensure logical consistency and high accuracy, making it particularly effective for mathematical, symbolic, or algorithmic tasks. PAL relies on a suitable code execution environment and consistently produces results with high reliability and stable, moderately sized outputs. However, it may not well suited for commonsense reasoning tasks.

\end{tcolorbox}

\vspace{-2cm}

\begin{tcolorbox}
\textbf{Qwen2.5-3B-Instruct:} Qwen2.5-3B-Instruct is a lightweight 3B parameter model with fast inference and low resource usage. It is suitable for simple tasks such as basic question answering and short-form text generation, but is limited in handling complex reasoning or multi-step tasks.\\

\textbf{Qwen2.5-7B-Instruct:} Qwen2.5-7B-Instruct is a mid-small 7B parameter model that balances speed and performance. It is capable of multi-turn dialogue, basic code and math tasks, and offers improved language understanding over smaller models, while maintaining efficient inference.\\

\textbf{Qwen2.5-14B-Instruct:} Qwen2.5-14B-Instruct is a mid-sized 14B parameter model that excels at complex reasoning, document summarization, and structured mid-length text generation. It demonstrates strong performance on tasks requiring deeper understanding and context retention.\\

\textbf{DeepSeek-R1-7B:} DeepSeek-R1-7B is a 7B distilled model with slightly slower but stable inference compared to other models of similar size. It is well-suited for medium-length answers that require deep reasoning, and often generates more detailed and comprehensive responses.\\

\textbf{DeepSeek-R1-14B:} DeepSeek-R1-14B is a 14B distilled model with slower inference but strong logical and mathematical capabilities. It is ideal for mathematical proofs and tasks requiring rigorous step-by-step reasoning, offering robust performance in logic-intensive scenarios.\\

\textbf{QwQ-32B:} QwQ-32B is a large 32B quantized model with slow inference. It excels at complex logic, coding, and multi-step reasoning tasks, though it may produce verbose outputs. Its large capacity enables handling of challenging prompts and long-context tasks.\\

\textbf{Qwen3-4B (thinking):} Qwen3-4B in 'thinking' mode generates longer reasoning chains and detailed thought processes. While it has slower inference and higher resource usage, it excels at solving complex logic and reasoning problems, making it suitable for tasks that require step-by-step explanations or in-depth analysis.\\

\textbf{Qwen3-4B (non-thinking):} Qwen3-4B in 'non-thinking' mode is optimized for short, direct answers. It provides fast inference with low resource cost, but is limited in deep reasoning or step-by-step explanations, making it best for straightforward queries or when efficiency is prioritized.
\end{tcolorbox}

%% file: cases/case1.tex
\begin{figure*}[htbp]
\centering
\begin{tcolorbox}[size=title,opacityfill=0.1]

\scriptsize
\setlength{\tabcolsep}{3pt}
\centering
Routing Table\\
\begin{tabular}{lccccc|cc}
    \toprule
    Question & Model & Strategy & Pred Perf & Pred Ans Length & Route Score & GT Label & GT Ans Length \\
    \midrule
    \multirow{14}{*}{\parbox{1.5cm}{\centering Raymond and Samantha are cousins. Raymond was \dots}} 
    & Qwen2.5-14B & CoT & 0.982 & 275.9 & 0.975 & True & 170.0 \\
    & QwQ-32B & CoD & 0.986 & 568.6 & 0.972 & True & 392.0 \\
    & QwQ-32B & CoT & 0.991 & 1131.5 & 0.963 & True & 635.0 \\
    & DeepSeek-R1-14B & CoD & 0.961 & 320.1 & 0.953 & True & 217.0 \\
    & DeepSeek-R1-14B & CoT & 0.971 & 789.2 & 0.951 & True & 290.0 \\
    & DeepSeek-R1-14B & Vanilla & 0.969 & 771.6 & 0.950 & True & 310.0 \\
    & Qwen2.5-7B & CoT & 0.948 & 310.9 & 0.940 & False & 229.0 \\
    & QwQ-32B & Vanilla & 0.979 & 1607.1 & 0.939 & True & 923.0 \\
    & Qwen2.5-14B & PAL & 0.927 & 209.2 & 0.921 & True & 157.0 \\
    & Qwen2.5-3B & CoT & 0.930 & 375.7 & 0.920 & False & 191.0 \\
    & DeepSeek-R1-7B & CoT & 0.946 & 1023.7 & 0.920 & False & 378.0 \\
    & DeepSeek-R1-7B & Vanilla & 0.945 & 1006.9 & 0.919 & False & 290.0 \\
    & DeepSeek-R1-7B & CoD & 0.917 & 299.3 & 0.910 & False & 144.0 \\
    \multicolumn{8}{c}{\dots} \\
    \bottomrule
\end{tabular}

\vspace{1em}

\normalsize
\raggedright
\noindent

{\color[HTML]{1E90FF} \faQuestionCircle\ \textit{Question (GSM8K)} }:\\
Raymond and Samantha are cousins. Raymond was born 6 years before Samantha. Raymond had a son at the age of 23. If Samantha is now 31, how many years ago was Raymond's son born? (\textbf{Ground Truth:} 14) \\
\vspace{1em}

{\color[HTML]{D72638} \faCog\ \textit{Fixed Best} $\xrightarrow{}$ \textit{QwQ-32B + CoT} }:\\
\texttt{<think>}
Okay, let's see. The problem is about figuring out how many years ago Raymond's son was born, given some information about their ages. Let me break it down step by step.
First, the question states that Raymond and Samantha are cousins, and Raymond was born 6 years before Samantha. So, Raymond is older than Samantha by 6 years. Currently, Samantha is 31 years old. That means Raymond's current age should be Samantha's age plus those 6 years. Let me check that again. Wait, if Raymond was born 6 years before Samantha, then yes, he is 6 years older. So if Samantha is 31 now, Raymond must be $31 + 6 = 37$ years old right now.
Next part: Raymond had a son at the age of 23. So, we need to find out how many years ago that was. Let's think. If Raymond is currently 37, and he was 23 when his son was born, then the number of years that have passed since his son was born is $37 - 23 = 14$. Wait, that would mean his son is 14 years old now. So the son was born 14 years ago. Hmm, but let me make sure I didn't mix up anything.
Wait, let me rephrase all the steps to confirm:
\textbf{(More Thinking Steps \dots)}
\texttt{</think>}\\

Raymond is 6 years older than Samantha. Since Samantha is currently 31, Raymond is $31 + 6 = 37$ years old. Raymond had his son at age 23, so the number of years since his son's birth is $37 - 23 = 14$.
The answer is $\boxed{14}$. \textbf{(Total 635 tokens)}

\vspace{1em}

{\color[HTML]{2EA121} \faLightbulbO\ \textit{RTR} $\xrightarrow{\text{route}}$ \textit{Qwen2.5-14B + CoT} }:\\
To solve this problem, we need to determine the current age of Raymond and then figure out how long ago his son was born. First, determine Raymond's current age. Samantha is currently 31 years old. Raymond was born 6 years before Samantha. Therefore, Raymond is currently 31 + 6 = 37 years old.
Next, determine when Raymond's son was born. Raymond had a son when he was 23 years old. Raymond is currently 37 years old. The difference between Raymond's current age and the age when he had his son is 37 - 23 = 14 years. Therefore, Raymond's son was born 14 years ago. \boxed{14}. \textbf{(Total 170 tokens)}

\end{tcolorbox}
\caption{An easy case from GSM8K. The fixed-best model-strategy pair incurs a token cost of 635. RTR predicts a high cost for this option and instead routes the question to a lightweight non-reasoning model (Qwen2.5-14B), which answers correctly using only 170 tokens.}
\label{fig:case1}
\end{figure*}

%% file: cases/case2.tex
\begin{figure*}[htbp]
\centering
\begin{tcolorbox}[size=title,opacityfill=0.1]

\scriptsize
\setlength{\tabcolsep}{3pt}
\centering
Routing Table\\
\begin{tabular}{lccccc|cc}
    \toprule
    Question & Model & Strategy & Pred Perf & Pred Ans Length & Route Score & GT Label & GT Ans Length \\
    \midrule
    \multirow{11}{*}{\parbox{1.5cm}{\centering Divide the reciprocal of $\frac{5}{6}$ by the the reciprocal of $\frac{5}{6}$.}} 
    & DeepSeek-R1-14B & CoD & 0.979 & 619.3 & 0.964 & True & 216.0 \\
    & QwQ-32B & CoD & 0.994 & 1414.0 & 0.959 & True & 1756.0 \\
    & Qwen2.5-14B & CoT & 0.967 & 453.3 & 0.956 & True & 333.0 \\
    & QwQ-32B & CoT & 0.995 & 2051.1 & 0.943 & True & 1963.0 \\
    & DeepSeek-R1-14B & CoT & 0.980 & 1815.3 & 0.934 & True & 3387.0 \\
    & DeepSeek-R1-7B & CoD & 0.946 & 720.0 & 0.928 & True & 196.0 \\
    & DeepSeek-R1-14B & Vanilla & 0.976 & 1978.9 & 0.927 & True & 2273.0 \\
    & QwQ-32B & Vanilla & 0.988 & 2726.9 & 0.920 & True & 2834.0 \\
    & Qwen2.5-7B & CoT & 0.922 & 468.6 & 0.910 & True & 400.0 \\
    & DeepSeek-R1-7B & CoT & 0.957 & 2367.1 & 0.898 & False & 17000.0 \\
    & DeepSeek-R1-7B & Vanilla & 0.954 & 2493.2 & 0.892 & True & 1972.0 \\
    \multicolumn{8}{c}{\dots} \\
    \bottomrule
\end{tabular}

\vspace{1em}

\normalsize
\raggedright
\noindent

{\color[HTML]{1E90FF} \faQuestionCircle\ \textit{Question (Math)} }:\\
Divide the reciprocal of $\frac{5}{6}$ by the reciprocal of the reciprocal of $\frac{5}{6}$. \textbf{(Ground Truth: $\frac{36}{25}$ )}\\
\vspace{1em}

{\color[HTML]{D72638} \faCog\ \textit{Fixed Best} $\xrightarrow{}$ \textit{QwQ-32B + CoT} }:\\
\texttt{<think>}
Okay, let's tackle this problem step by step. The question says: "Divide the reciprocal of 5/6 by the reciprocal of the reciprocal of 5/6." Hmm, that sounds a bit convoluted, but let's break it down.
First, I need to recall what a reciprocal is. The reciprocal of a fraction is just flipping the numerator and denominator. So, the reciprocal of 5/6 would be 6/5. Got that part.
Now, the problem mentions "the reciprocal of the reciprocal of 5/6." Let me parse that. The first reciprocal of 5/6 is 6/5. Then the reciprocal of that would be flipping 6/5 back to 5/6. So, the reciprocal of the reciprocal of 5/6 is just 5/6 again. Wait, so the second part simplifies back to the original fraction? That makes sense because taking the reciprocal twice cancels out the operation.
So now the problem is asking me to divide the first reciprocal (which is 6/5) by the second term, which we just found out is 5/6. So mathematically,
\textbf{(More Thinking Steps \dots)}
\texttt{</think>}\\

First, we need to find the reciprocal of $\frac{5}{6}$, which is $\frac{6}{5}$.
Next, we determine the reciprocal of the reciprocal of $\frac{5}{6}$. The reciprocal of $\frac{6}{5}$ is $\frac{5}{6}$.
We then need to divide the first result by the second result. This means we need to compute $\frac{6}{5} \div \frac{5}{6}$.
To divide by a fraction, we multiply by its reciprocal:
\[
\frac{6}{5} \div \frac{5}{6} = \frac{6}{5} \times \frac{6}{5} = \frac{6 \times 6}{5 \times 5} = \frac{36}{25}
\]

Thus, the final answer is $\boxed{\dfrac{36}{25}}$. \textbf{(Total 1963 tokens)}

\vspace{1em}

{\color[HTML]{2EA121} \faLightbulbO\ \textit{RTR} $\xrightarrow{\text{route}}$ \textit{DeepSeek-R1-14B + CoD} }:\\
\texttt{<think>}
First, I need to find the reciprocal of 5/6, which is 6/5.
Next, I determine the reciprocal of the reciprocal of 5/6. The reciprocal of 5/6 is 6/5, and the reciprocal of that is 5/6.
Finally, I divide the first reciprocal (6/5) by the second reciprocal (5/6), which simplifies to (6/5) * (6/5) = 36/25.
\texttt{</think>}\\
Step-by-step minimal draft:
\begin{enumerate}
    \item Reciprocal of 5/6: 6/5
    \item Reciprocal of reciprocal: 5/6
    \item Divide: (6/5) $\div$ (5/6)
    \item Multiply by reciprocal: (6/5) $\times$ (6/5)
    \item Result: 36/25
\end{enumerate}
The Answer is $\boxed{\dfrac{36}{25}}$. \textbf{(216 tokens)}

\end{tcolorbox}
\caption{A case from Math. The fixed-best model-strategy pair answers the question with a token cost of 1963. RTR instead selects DeepSeek-R1-14B with Chain-of-Thought (CoD), which correctly solves the problem using only 216 tokens. Since the question benefits from a step-by-step draft-style reasoning process, this pair provides a clear and efficient solution at a much lower token cost.}
\label{fig:case2}
\end{figure*}

%% file: cases/case3.tex
\begin{figure*}[htbp]
\centering
\begin{tcolorbox}[size=title,opacityfill=0.1]

\scriptsize
\setlength{\tabcolsep}{3pt}
\centering
Routing Table\\
\begin{tabular}{lccccc|cc}
    \toprule
    Question & Model & Strategy & Pred Perf & Pred Ans Length & Route Score & GT Label & GT Ans Length \\
    \midrule
    \multirow{21}{*}{\parbox{1.5cm}{\centering For any two different real numbers $x$ and $y$\dots}} 
    & QwQ-32B & CoD & 0.556 & 9756.9 & 0.312 & True & 9157.0 \\
    & Qwen2.5-14B & CoT & 0.322 & 898.0 & 0.300 & False & 599.0 \\
    & Qwen2.5-14B & PAL & 0.308 & 580.2 & 0.294 & False & 678.0 \\
    & QwQ-32B & CoT & 0.569 & 11623.6 & 0.278 & True & 11767.0 \\
    & Qwen2.5-14B & CoD & 0.273 & 190.8 & 0.268 & False & 90.0 \\
    & Qwen2.5-14B & Vanilla & 0.280 & 935.0 & 0.257 & False & 518.0 \\
    & Qwen2.5-7B & Vanilla & 0.246 & 915.2 & 0.223 & False & 695.0 \\
    & QwQ-32B & Vanilla & 0.519 & 12125.3 & 0.215 & True & 11671.0 \\
    & Qwen2.5-7B & CoD & 0.209 & 256.5 & 0.202 & False & 73.0 \\
    \multicolumn{8}{c}{\dots} \\
    & DeepSeek-R1-14B & CoD & 0.391 & 10674.2 & 0.124 & True & 3422.0 \\
    & Qwen2.5-3B & CoD & 0.113 & 384.0 & 0.104 & False & 170.0 \\
    & Qwen2.5-3B & CoT & 0.131 & 1119.9 & 0.103 & False & 621.0 \\
    & DeepSeek-R1-7B & CoD & 0.289 & 9690.2 & 0.046 & False & 3762.0 \\
    & DeepSeek-R1-7B & CoT & 0.342 & 12369.1 & 0.033 & False & 17000.0 \\
    & DeepSeek-R1-14B & CoT & 0.336 & 13275.0 & 0.004 & False & 20481.0 \\
    & DeepSeek-R1-7B & Vanilla & 0.308 & 12749.7 & -0.011 & False & 17000.0 \\
    & DeepSeek-R1-14B & Vanilla & 0.320 & 13563.4 & -0.019 & False & 20481.0 \\
    \bottomrule
\end{tabular}

\vspace{1em}

\normalsize
\raggedright
\noindent

{\color[HTML]{1E90FF} \faQuestionCircle\ \textit{Question (OlympiadBench)} }:\\
For any two different real numbers $x$ and $y$, we define $D(x, y)$ to be the unique integer $d$ satisfying $2^{d} \leqslant |x-y|<2^{d+1}$. Given a set of reals $\mathcal{F}$, and an element $x \in \mathcal{F}$, we say that the scales of $x$ in $\mathcal{F}$ are the values of $D(x, y)$ for $y \in \mathcal{F}$ with $x \neq y$. Let $k$ be a given positive integer. Suppose that each member $x$ of $\mathcal{F}$ has at most $k$ different scales in $\mathcal{F}$ (note that these scales may depend on $x$ ). What is the maximum possible size of $\mathcal{F}$ ? \textbf{(Ground Truth: \( 2^k \))}\\
\vspace{1em}

{\color[HTML]{D72638} \faCog\ \textit{Fixed Best} $\xrightarrow{}$ \textit{QwQ-32B + CoT} }:\\
\texttt{<think>}Okay, so I need to figure out the maximum possible size of a set $F$ where each element in $F$ has at most $k$ different scales. The scales here are defined as the values of $D(x, y)$ for each pair of distinct elements $x$ and $y$ in $F$. And $D(x, y)$ is the unique integer $d$ such that $2^d \leq |x - y| < 2^{d+1}$.
Hmm, let me start by understanding the problem step by step.
\textbf{(More Thinking Steps \dots)}
\texttt{</think>}\\
To solve the problem, we need to determine the maximum possible size of a set $\mathcal{F}$ such that each element in $\mathcal{F}$ has at most $k$ different scales. The scales are defined by the function $D(x, y)$, which categorizes the distance between any two elements $x$ and $y$ into intervals of powers of two.
\textbf{(More Reasoning Steps \dots)}\\
The recursive doubling of the set size for each increment in $k$ leads to the conclusion that the maximum size of $\mathcal{F}$ is $2^k$. The maximum possible size of the set $\mathcal{F}$ is $\boxed{2^k}$.
\textbf{(11767 tokens)}
\vspace{1em}

{\color[HTML]{2EA121} \faLightbulbO\ \textit{RTR} $\xrightarrow{\text{route}}$ \textit{QwQ-32B + CoD} }:\\
\texttt{<think>}Okay, so I need to figure out the maximum possible size of a set $F$ where each element has at most $k$ different scales. The scales are defined as $D(x, y)$, which is the integer $d$ such that $2^d \leq |x - y| < 2^{d+1}$. Each element $x$ in $F$ can have at most $k$ different $D(x, y)$ values when compared to all other elements in $F$.
Hmm, let's start by understanding the problem better.
\textbf{(More Thinking Steps \dots)}
\texttt{</think>}\\

\begin{enumerate}
    \item Define scales via \( D(x, y) = d \) where \( 2^d \leq |x - y| < 2^{d+1} \).
    \item Each element \( x \) has at most \( k \) different \( D(x, y) \) values.
    \item For maximum size, arrange points such that pairwise distances fall into \( k \) scales.
    \item Construct the set \( \{0, 1, 2, \ldots, 2^k - 1\} \): each element's distances to others fall into scales \( 0 \) to \( k - 1 \).
    \item This gives exactly \( k \) scales per element, achieving \( 2^k \) elements.
\end{enumerate}

The Answer is $\boxed{2^k}$
\textbf{(9157 tokens)}

\end{tcolorbox}
\caption{A hard question from OlympiadBench. As shown in the routing table, most model-strategy pairs fail to answer this question. RTR accurately selects the strongest model, QwQ-32B, along with the CoD strategy, which reduces the token cost while preserving answer correctness. This leads to a correct answer with a token reduction of over 2500.}
\label{fig:case3}
\end{figure*}